\documentclass[10pt, a4paper]{article}

\usepackage{lrec-coling2024} 
\usepackage{arydshln}

\title{Two-step Automated Cybercrime Coded Word Detection \\ using Multi-level Representation Learning}

\name{Yongyeon Kim$^1$, Byung-Won On$^{1*}$\thanks{$^*$Corresponding author}, Ingyu Lee$^2$} 

\address{School of Computer Science and Engineering$^1$, Institute of Information and Communication$^2$ \\
         Kunsan National University$^1$, Yeungnam University$^2$ \\
         dyddus1210@kunsan.ac.kr, bwon@kunsan.ac.kr, inlee3471@gmail.com\\}

\abstract{
In social network service platforms, crime suspects are likely to use cybercrime coded words for communication by adding criminal meanings to existing words or replacing them with similar words.
For instance, the word `ice' is often used to mean methamphetamine in drug crimes. To analyze the nature of cybercrime and the behavior of criminals, quickly detecting such words and further understanding their meaning are critical. In the automated cybercrime coded word detection problem, it is difficult to collect a sufficient amount of training data for supervised learning and to directly apply language models that utilize context information to better understand natural language. To overcome these limitations, we propose a new two-step approach, in which a mean latent vector is constructed for each cybercrime through one of five different AutoEncoder models in the first step, and cybercrime coded words are detected based on multi-level latent representations in the second step. Moreover, to deeply understand cybercrime coded words detected through the two-step approach, we propose three novel methods: (1) Detection of new words recently coined, (2) Detection of words frequently appeared in both drug and sex crimes, and (3) Automatic generation of word taxonomy. According to our experimental results, among various AutoEncoder models, the stacked AutoEncoder model shows the best performance. Additionally, the F1-score of the two-step approach is 0.991, which is higher than 0.987 and 0.903 of the existing dark-GloVe and dark-BERT models. By analyzing the experimental results of the three proposed methods, we can gain a deeper understanding of drug and sex crimes.
 \\ \newline  }

\begin{document}

\maketitleabstract
\section{Introduction}
With the advent of mobile devices and the rapid development of Internet technology, Social Network Service (SNS) platforms such as X (Twitter), Telegram, Meta Threads, and Reddit have emerged. This allows people to freely exchange their opinions, creating a huge amount of data in real time. Through this data explosion, users can obtain high-quality information, but they are inevitably exposed to serious cybercrimes.

~\citet{GMMcyberbullying2020} has reported that about 43\% of teenagers in the United States have been victims on social media.
In a survey of 10,008 teenagers aged 13 to 22 conducted by Ditchthelabel and Habbo~\cite{LSHWE2020}, approximately 37\% of teenagers suffered from frequent cybercrimes.
In Korea, drug crimes due to drug advertising and delivery through SNS increased by 13.9\% in 2022 compared to the previous year~\cite{bkim-22}, and despite the punishment of sex crimes being strengthened due to the Telegram n-room incident, digital sex crimes increased by 37.2\% in 2022 compared to the previous year~\cite{police-22}. As such, cybercrime is constantly increasing.

Accordingly, various types of words used in cybercrime are emerging. Cybercrime terms is used to secretly share information and avoid the surveillance of investigators. Therefore, the argot used by cybercriminals is important information in analyzing crime characteristics and behavior. In other words, identifying lingo used in crime and detecting new crime terms as quickly as possible is very important in preventing cybercrime.

\begin{table*}
\centering
\begin{tabular}{lll}
\hline
\textbf{Category} & \textbf{Sentence} & \textbf{Criminal meaning}\\
\hline
Drugs & ``we sell clean \textbf{ice}'' & Methamphetamine \\
      & ``i am looking for \textbf{santa claus}'' & Drug supplier \\
\hline
Sex crimes & ``contact me if you want to buy a \textbf{mv}'' & Masturbation video \\
           & ``looking for a high-paying \textbf{sek part-time job}'' & Prostitution \\
\hline
\end{tabular}
\caption{\label{slangs} Examples of cybercrime coded words.}
\end{table*}

As shown in Table~\ref{slangs}, a cybercrime coded word is commonly used by adding criminal meaning to existing general words. For example, in the sentence ``i am looking for santa claus,'' the word `santa claus' refers to a drug supplier. As another example, in the sentence ``we sell clean ice,'' the word `ice' is used to mean methamphetamine in drug crimes. Sometimes the word `cold' is used instead of ice. Since these cybercrime words are very volatile, they are often used within a short period of time and soon disappear to avoid police crackdown. In this study, we will formally name such a cybercrime term a \textbf{cybercrime coded word}~\footnote{We will use this term frequently throughout this paper. In addition, we will often abbreviate it as a C3 word.} that indicates an ordinary-sounding word with a secret meaning to conceal what they are discussing~\footnote{In this problem, the word `ice' in Table~\ref{slangs} is a true positive example. On the other hand, actual drug names or slurs (e.g., `ecstasy', `whore', and `pussy') are false positive examples.}. This is, these terms refer to encrypted terms that cybercriminals use in the process of exchanging information, circumventing security and make it difficult for law enforcement or security experts to understand. Especially, we define this study as the problem of automatically detecting new C3 words and further understanding their meaning, given a large number of documents related to cybercrime such as drugs or sex crimes.

However, identifying new C3 words in cybercime has two limitations. In general, cybercrime vocabulary is likely to be used for a while and then replaced by another term soon because cybercrime occurs secretly, so it is difficult to collect a sufficient amount of training data for supervised learning. In addition, because it is common for criminal meanings to be added to words with their original meaning, it is difficult to directly apply a language model such as Transformer or BERT that utilizes context information to better understand natural language.

To overcome these limitations, we focus on semi-supervised learning for cybercrime coded word detection. In particular,
we propose a new \textbf{two-step approach} based on AutoEncoder framework that consists of Stacked AutoEncoder (SAE), Adversarial AutoEncoder (AAE), Denoising AutoEncoder (DAE), Variational AutoEncoder (VAE), and Stacked Denoising AutoEncoder (SDAE). In the first step, we construct the mean latent vector $\bar{v_{c_i}}$ by the $i$-th cybercrime type through one of the AutoEncoder models with a small amount of training data set in cybercrime. For example, the mean latent vectors $\bar{v_{c_1}}$ and $\bar{v_{c_2}}$, where $c_1$ and $c_2$ indicate drugs and sex crimes, respectively.

In the second step, the proposed model detects C3 words using multi-level representations which consist of the sentence-level and word-level ones. Specifically, given a new document $d$ of interest, $d$ is divided into a set of sentences $s_i \in \{s_1, s_2, ..., s_m\}$, where $m$ is the number of sentences in $d$. Each $s_i$ is encoded to the latent vector $\vec{v_{s_i}}$ by one of the AutoEncoder models. If $sim(\vec{v_{s_i}}, \bar{v_{c_1}}) > sim(\vec{v_{s_i}}, \bar{v_{c_2}})$, $\vec{v_{s_i}}$ is labelled to $c_1$, and $c_2$, otherwise. In the word-level representation, each word $w_j \in s_i$ is encoded to $\vec{v_{w_j}}$, similar to the sentence-level representation process. If $sim(\vec{v_{w_j}}, \bar{v_{c_1}}) > \theta$ and $w_j \notin D_{c_1}, \textnormal{the C3 word dictionary to } c_1$, $w_j$ is considered to be a C3 word for $c_1$.

In addition, to deeply understand C3 words detected through our two-step approach, we propose three novel methods. The first method is to find new, unknown C3 words rather than 
to find already well-known ones in cybercrime. The second method is to detect criminal posts on the SNS platforms that may contain C3 words across two or more cybercrimes. The third method is to automatically generate the taxonomy of C3 words detected by the two-step approach. Our study of these methods is the first of its kind, as far as we are aware.

Our experimental results show that the SAE-based two-step approach is the best, compared to the others. In addition, its F1-score is 0.991, which is higher than 0.987 and 0.903 of dark-GloVe and dark-BERT models that are known as the state-of-the-art models in C3 word detection. Moreover, through our in-depth analysis of the detected words, we can understand new C3 words recently coined, C3 words frequently appeared in both drug and sex crimes, and the taxonomy of C3 words. Please take a look at the details in the experimental section.

In this work, our technical contributions are as follows.
\begin{itemize}
\item To detect cybercrime coded words, we propose a novel two-step approach, in which the mean latent vector is computed by type of cybercrime through the best of five different AutoEncoder models in the first step, and cybercrime coded words are detected based on multi-level latent representations in the second step. 
\item To deeply understand cybercrime coded words detected through the two-step approach, we propose three methods: (1) Detection of new words recently coined, (2) Detection of words frequently appeared in both drug and sex crimes, and (3) Automatic generation of cybercrime coded word taxonomy.
\item Our experimental results show that the F1-score of the two-step approach is 0.991, which is higher than 0.987 and 0.903 of the baseline models such as dark-GloVe and dark-BERT models. In addition, we can gain a deeper understanding of drug and sex crimes by analyzing the cybercrime coded words detected through the two-step approach in three aspects.
\end{itemize}

\section{Related Works}
\label{rel}

Traditionally, supervised learning approach has been used to detect cybercrime coded words including cyberbullying and illegal web pages in the Internet. \citet{Morris2012IdentifyingSP} applied SVM classifier to identify sexual predators in chatting message systems considering lexical features such as emoticons with behavioral features. \citet{Latapy2012QuantifyingPA} investigated peer-to-peer networks with predefined list of keywords to detect child-pornography queries. \citet{al-nabki-etal-2017-classifying} manually classified web pages into topical categories and subcategories and provided the data set DUTA (Darknet Usage Text Addresses). Later, they extended the data set to DUTA-10k~\cite{al-nabki-etal-2019}. \citet{Avarikioti2018StructureAC} presented topic classification of text into legal and illegal activities using a SVM classifier with bag-of-words features. 
 
To automatically detect cyberbullying, \citet{cyberbullying2017} proposed semantic-enhanced marginalized denoising AutoEncoder (smSDA) using dropout noise and sparsity constraints. The dropout noise is designed considering domain knowledge and word embedding technique. 
\citet{GMMcyberbullying2020} proposed an unsupervised cyberbullying detection via time-informed Gaussian Mixture Model that simultaneously fits the comment inter-arrival times and estimates the bullying likelihood. The authors encode the social media session by exploiting multi-modal features including text, network and time.

Word embedding methods cannot represent well on deliberately obfuscated words which are generally considered as rare words with a little contextual information and generally removed during preprocessing phase. \citet{LSHWE2020} proposed a model improving similarity-based word embedding with locality sensitive hashing to detect cyberbullying. They exploited the facts that the deliberately obfuscated words have a high context similarity with their corresponding bullying words.
\citet{euphemism2021} analyzed words in sentence-level context to detect words being used euphemistically which showed higher accuracy compared to word embedding approaches.

\citet{choshen-etal-2019-language} explored the differences in the legal and illegal pages on the Dark Web. They utilized the distinguishing features between the two domains including named entity, vocabulary, and syntactic structure to detect illegal pages using standard NLP tools with considerable domain adaptation.
\citet{jin-etal-2022-shedding} also noticed the linguistic differences in features such as POS distributions and vocabulary usages  between Surface Internet and Dark Web, and conjectured that pretrained language models in Dark Web performs better than simple lexical models with domain specific pretrained language model. They also proposed the DarkBERT~\citep{jin-etal-2023-darkbert} which is a language model pretrained on Dark Web data set. According to the authors, DarkBERT is better suited for NLP tasks on Dark Web specific text.
\citet{ranaldi2022dark} also explored the use of pretrained language models over Dark Web text. They claimed that lexical and syntactic models such as GloVe~\citep{pennington-etal-2014-glove} after training domain-specific data outperforms pretrained language model such as BERT~\citep{devlin-etal-2019-bert}.

Many researches have been done to detect C3 words including cyberbullying and cybercrime.
However, the aforementioned approaches require time consuming and labor intensive data labelling process. In addition, current labeling guidelines may not be suitable in the future with the changes of language usage or appearance of new C3 words.
As authors aware, this is the first approach to automatically detect and build C3 words on Surface and Dark Internet environments.

\section{Proposed Model}
\begin{figure*}[!ht]
\begin{center}
\includegraphics[scale=0.53]{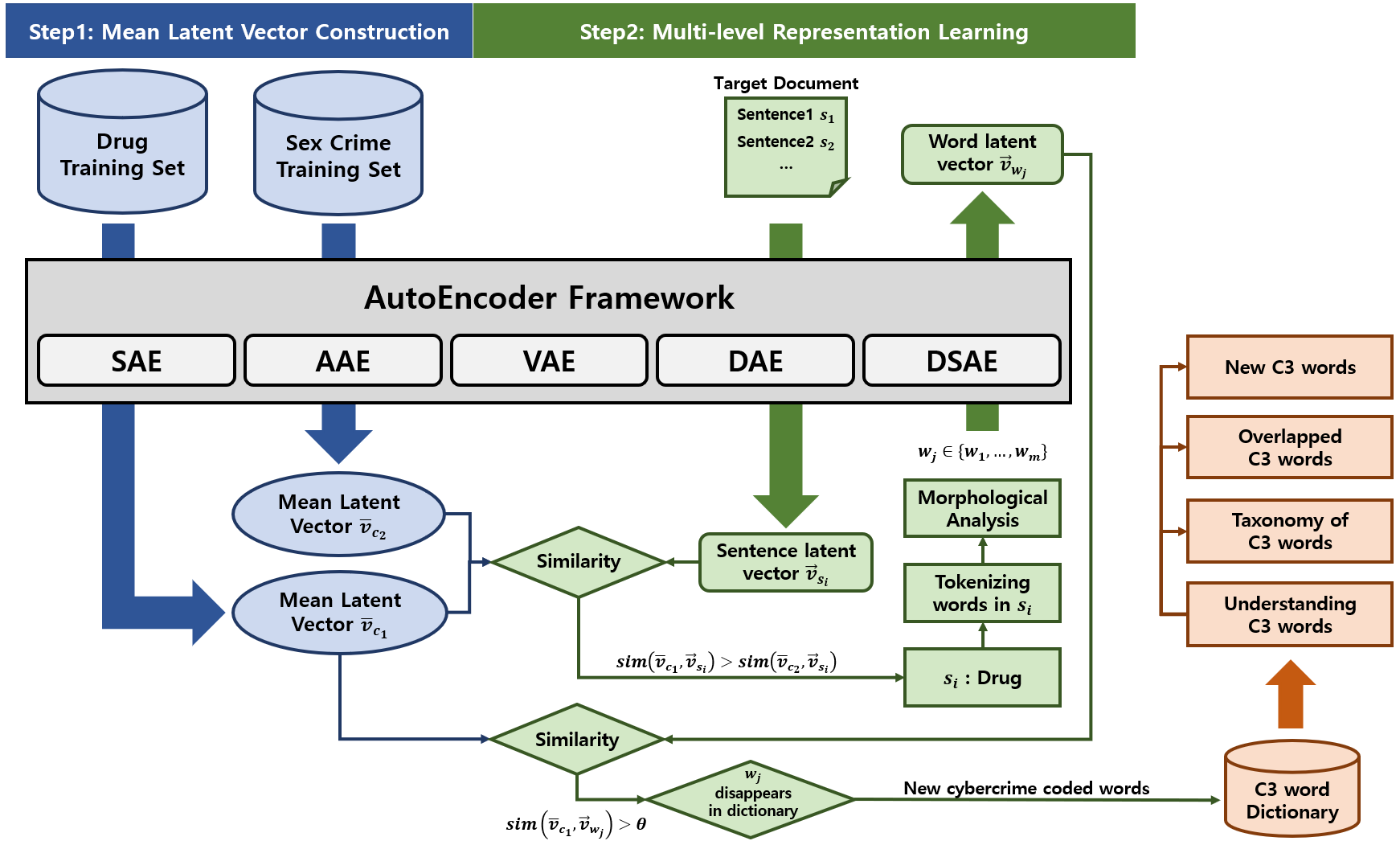} 
\caption{Overview of the proposed two-step approach.}
\label{fig:sysoverview}
\end{center}
\end{figure*}

To address the cybercrime coded word detection problem, we propose two-step approach, as illustrated in Figure~\ref{fig:sysoverview}. The first step is to obtain the mean latent vector for each type of cybercrimes, such as drugs or sex crimes, from a small amount of training data. In Section~\ref{step1}, we will describe the details. The second step is to detect candidates that may signify new C3 words for a certain cybercrime, given a target document. In Section~\ref{step2}, we will describe the second step in detail.

\begin{figure}[!ht]
\begin{center}
\includegraphics[scale=0.57]{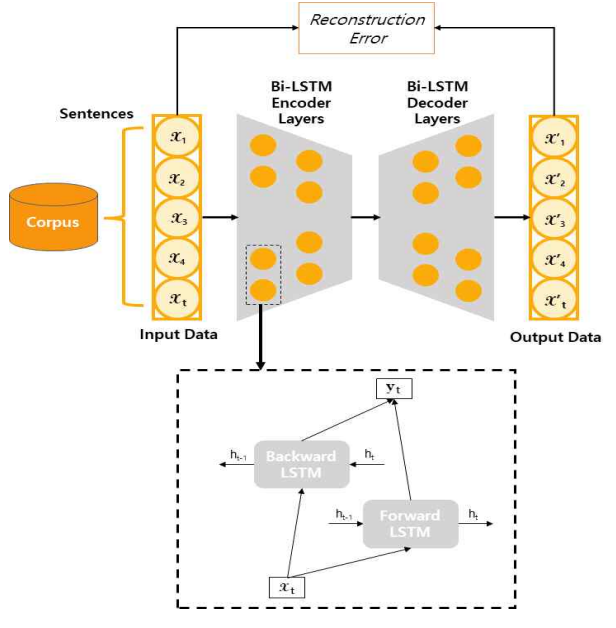} 
\caption{The proposed AutoEncoder model based on Bi-LSTM.}
\label{autoencoder}
\end{center}
\end{figure}

\subsection{Step 1: Construction of Mean Latent Vector by Type of Cybercrime}
\label{step1}

Suppose a set of cybercrime classes $C=\{\textnormal{`drugs'}, \textnormal{`sex crimes'}\}$ and a small amount of training data set $T_c$ containing documents related to $c \in C$. The goal of this step is to compute the mean latent vector $\bar{v_c}$ for $c \in C$. Firstly, $T_c$ is divided into the training and test sets, $T_c^{tr}$ and $T_c^{ts}$. Next, we map the $i$-th document $d_i$ in $T_c^{tr}$ to the low-dimensional latent vector $\vec{v_i}$ through an autoencoder model $f$.

\begin{equation}
\vec{v_i} = f(d_i)
\label{encoding}
\end{equation}

For $f$, we consider five different AutoEncoder models based on Bi-LSTM, which are Adversarial AutoEncoder (AAE), Denoising AutoEncoder (DAE), Stacked AutoEncoder (SAE), Stacked Denoising AutoEncoder (SDAE), and Variational AutoEncoder (VAE). Each AutoEncoder model maps each document related to cybercrime in $T$ to a low-dimensional vector. As shown in Figure~\ref{autoencoder}, because the encoder and decoder of the proposed model are composed of bidirectional LSTM cells, our model works well in encoding and compressing the document with variable length and dependent relationships between words, such as text sequences, into latent vectors. This AutoEncoder model is trained by performing error backpropagation to minimize reconstruction error. The latent vectors generated through the AutoEncoder model contain important information from the original data by preserving meaningful information in projecting the latent space. Finally, we obtain the mean latent vector $\bar{v_c}$ for $c \in C$ from $T_c^{ts}$ through Equation~\ref{meanv}.

\begin{equation}
\bar{v_c} = \frac{1}{N}\Sigma_{i=1}^n \vec{v_i}
\label{meanv}
\end{equation}
, where $N$ is the number of latent vectors in $T_c^{ts}$.

\subsubsection{Stacked AutoEncoder (SAE)}
\label{subsubsec:sae}
A stacked AutoEncoder is a deep neural network mainly used in unsupervised learning and has been widely used in research to understand the implied meaning of text. In general, it consists of an encoder and a decoder. The encoder receives text data and gradually reduces its dimensionality as it passes through several layers. In this process, important information from the input text is extracted and compressed into low-dimensional latent vectors. This latent vector represents the implied meaning and characteristics of the input text. Conversely, the decoder reconstructs the original text data using the latent vectors generated by encoding. The model is trained with the goal of minimizing the difference between input data and reconstructed data through a binary cross-entropy loss function $L_{SAE}$ as shown in Equation~\ref{eq:sae}. Therefore, the model with a smaller reconstruction error are likely to produce latent vectors containing more meaningful information of the original data.

\begin{equation}
\begin{array}{cl}
L_{SAE}(X, X^{'})
=-\frac{1}{N}\Sigma_{i=1}^n \{ X_i log(X_i^{'}) + \\ 
(1-X_i)log(1-X_i^{'})\}
\end{array}
\label{eq:sae}
\end{equation}
, where $X$ and $X^{'}$ mean the original text data and the reconstructed one.

\subsubsection{Denoising AutoEncoder (DAE)}
\label{subsubsec:dae}
A denoising AutoEncoder creates corrupted input data by adding some noise to text data. This is a modified model of a stacked AutoEncoder that reconstructs the impure input data back into the original noise-free text data. Four types of noise are used to add noise to the input test data. First, it shuffles the word order that makes up the sentence within the specified shuffling range. Second, words that make up the sequence are randomly removed according to a specified probability. Third, words are replaced by empty tokens according to the specified probability. Fourth, a word is replaced by another word according to the specified probability. By adding noise to the input data in this way, we can create a model that is robust to noise. It consists of an encoder and a decoder like the existing AutoEncoder model, and the encoder receives input data with added noise and generates a latent vector. Conversely, the decoder reconstructs the original noise-free text data from the latent vectors generated by the encoder. The model uses a loss function such as Equation~\ref{eq:sae} to remove noise as much as possible during the training step and restore the original text data.

\subsubsection{Stacked Denoising AutoEncoder (SDAE)}
A stacked denoising AutoEncoder is a model that combines SAE and DAE, which we already described in Sections~\ref{subsubsec:sae} and~\ref{subsubsec:dae}.~\citet{cyberbullying2017} proposed a cyberbullying detection model based on SDAE. However, they did not directly address cybercrime coded word detection in cybercrime in their study. 

\subsubsection{Variational AutoEncoder (VAE)}
A variational AutoEncoder is a model that learns latent vectors of data and uses them to generate new data. Unlike stacked AutoEncoder, it learns the probabilistic distribution of the latent space to generate new data that is similar but not identical to the input data. This is being used in research to understand the semantic structure embedded in text and generate text in latent space. It consists of an encoder that maps text data to a latent space. However, instead of creating a latent vector as a single value, two vectors, mean and variance, are created. These two vectors represent information about a given input text by defining a normal distribution in the latent space. The latent vectors sampled from the normal distribution $N(\mu, \sigma^2)$ are input to the decoder to generate data similar to the input data. The loss function used for model learning consists of two types. First, the reconstruction error is measured like the stacked AutoEncoder, and the method is as shown in Equation~\ref{eq:sae}. Second, it measures the difference between the distribution of the latent vector generated by inputting text data into the encoder and the normal distribution, and adds Kullback-Leibler (KL) divergence as in Equation~\ref{eq:vae} to make it similar to the normal distribution. In other words, by adding KL divergence to the reconstruction error loss function as shown in Equation~\ref{eq:vae2}, the latent vectors generated through the encoder are learned to resemble a normal distribution.

\begin{equation}
\begin{array}{cl}
L_{VAE}(X, X^{'}) = L_{SAE}(X, X^{'}) + \\ 
KL[N(\mu, \sigma^2)||N(0,1)]
\end{array}
\label{eq:vae}
\end{equation}

, where
\begin{equation}
\begin{array}{cl}
KL[N(\mu, \sigma^2)||N(0,1)]=-\frac{1}{2}(1+log \sigma - \mu^2 - \sigma^2)
\end{array}
\label{eq:vae2}
\end{equation}

\subsubsection{Adversarial AutoEncoder (AAE)}
An adversarial AutoEncoder is a model that integrates a variational encoder and Generative Adversarial Network (GAN). It consists of two different networks: a generator and a discriminator. The discriminator judges only the actual data distribution as true, and the generator creates a fake data distribution close to the real data distribution to prevent the discriminator from judging it. Different networks are trained adversarially to generate data similar to real data. In AAE, the generator role is performed by a differential AutoEncoder, and the discriminator role is performed by GAN. The generator receives input text data and samples latent vectors. The discriminator is responsible for determining whether a given latent vector is a fake encoded latent vector or a real latent vector sampled directly from a normal distribution. The adversarial AutoEncoder is trained in the following order: Encoder, Decoder, Discriminator, and Generator. The process of calculating the reconstruction error for input data $x$ as in Equation~\ref{eq:sae} and updating the encoder parameter $\phi$ and the decoder parameter $\theta$ is as in Equation~\ref{eq:aae1}. A discriminator is learned using the objective distribution $z_i$ and $q_{\phi}(x_i)$ sampled through the encoder, and the discriminator parameter $\lambda$ is updated as shown in Equation~\ref{eq:aae2}. Finally, learning progresses by updating $\phi$ through constructor learning, as shown in Equation~\ref{eq:aae3}.

\begin{equation}
L_i(\phi, \theta, x_i)=-E_{q_{\phi}(z|x_i)}[log(p_{\theta}(x_i|z))]
\label{eq:aae1}
\end{equation}

\begin{equation}
\begin{array}{cl}
-V(\phi, \lambda, x_i, z_i)=-log d_{\lambda}(z_i) - log(1-d_{\lambda}(q_{\phi}(x_i)))
\end{array}
\label{eq:aae2}
\end{equation}

\begin{equation}
-V(\phi, \lambda, x_i, z_i)=-log d_{\lambda}(q_{\phi}(x_i))
\label{eq:aae3}
\end{equation}

\begin{figure*}[!ht]
\begin{center}
\includegraphics[scale=0.55]{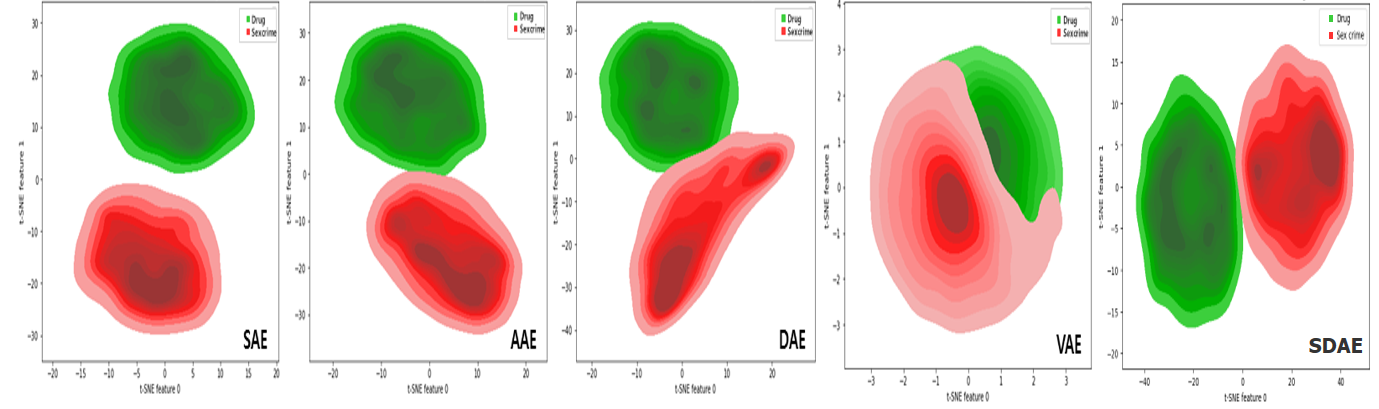} 
\caption{t-SNE visualization of latent vectors for C3 words detected through five AutoEncoder models.}
\label{fig.2}
\end{center}
\end{figure*}

\subsection{Step 2: Multi-level Latent Representations-based Cybercrime Coded Word Detection}
\label{step2}

Given a target document $d$ as input, the purpose of this step is to extract new cybercrime coded words from $d$. Our approach is based on
multi-level latent representations through $f$ in Section~\ref{step1}. One is the sentence-level latent representation and the other is the word-level one.

In the sentence-level latent representation, $d$ is segmented to a set of sentences. Through Equation~\ref{encoding}, each sentence is encoded to $\vec{v_i} = f(s_i)$, where $s_i$ is the $i$-th sentence in $d$. Then, a similarity score $sim$ is computed between $\vec{v_i}$ and each vector of the mean latent vector set $\{ \bar{v_{c_1}}, \bar{v_{c_2}} \}$, where $c_1$ and $c_2$ are drugs and sex crimes, respectively.

\begin{equation}
sim(\vec{v_i}, \bar{v_{c_j}}) = \frac{\vec{v_i} \cdot \bar{v_{c_j}}}{ \Vert \vec{v_i} \Vert \Vert \bar{v_{c_j}} \Vert } 
\end{equation}
, where $j \in \{1, 2\}$. If $sim(\vec{v_i}, \bar{v_{c_j}}) \geq \theta$, the sentence $s_i$ corresponding to $\vec{v_i}$ is classified to $c_j$. This is, the class of $s_i$ is $c_j$.

In the word-level latent representation, we first collect all sentences classified as $c_j$ and each sentence $s_i$ is tokenized to a set of word tokens. Through Equation~\ref{encoding}, each token is encoded to $\vec{v_{w_i}} = f(w_i)$, where $w_i$ is the $i$-th token in $s_i$. Then, a similarity score $sim$ is computed between $\vec{v_{w_i}}$ and each vector of the mean latent vector set $\{ \bar{v_{c_1}}, \bar{v_{c_2}} \}$, where $c_1$ and $c_2$ are drugs and sex crimes, respectively. If $sim(\vec{v_{w_i}}, \bar{v_{c_j}}) \geq \theta$, the token $w_i$ corresponding to $\vec{v_{w_i}}$ may be likely to be close to $c_j$. Furthermore, if $w_i$ does not appear to the existing dictionary $D$ of cybercrime coded words, we add $w_i$ to $D$ because we consider it to be a new C3 word.

\subsection{Analysis of C3 Words}

In this section, we propose three methods in order to deeply understand C3 words detected through our two-step approach.

\subsubsection{Detection of New C3 Words}
\label{sec:3.3.1}
The first proposed method is to find new, unknown words rather than to find already well-known words in cybercrime.
Figure~\ref{fig.2} shows the cluster results of latent vectors for C3 words detected through the five AutoEncoder models. The green is a cluster of C3 word vectors for drugs, and the red is a cluster of C3 word vectors for sexual crimes.

Please, take a look at the green. The green cluster is almost circular in shape. We hypothesize that the vectors at the center of the circle correspond to C3 words that are older in time, and the outlier vectors at the edges of the circle are new C3 words that were recently created. Therefore, the goal of the first method is to find outlier vectors. For this, we calculate the mean $\mu$ and standard deviation $\sigma$ of the latent vectors belonging to the green cluster. Once both $\mu$ and $\sigma$ for a group are known, a normal distribution $N(\mu, \sigma^2)$ can be approximated using Equation~\ref{nd}.

\begin{equation}
f(x) = \frac{1}{\sqrt{2 \pi} \sigma} e^{\frac{-(x- \mu^2}{(2\sigma^2)})}, -\infty <x < \infty
\label{nd}
\end{equation}

Then, we set the interval ($\hat{\theta_1}$, $\hat{\theta_2}$) as a 95\% confidence interval for parameter $\theta$, where $\hat{\theta_1}$ and $\hat{\theta_2}$ are called the low confidence limit and upper confidence limit of the confidence interval, respectively.
In this setting, since the population variance $\sigma$ is known in advance, the 95\% interval estimate for the population mean $\mu$ of $N(\mu, \sigma^2)$ is given in Equation~\ref{conf}.

\begin{equation}
(\bar{X}-z_{\frac{\alpha}{2}} \frac{\alpha}{\sqrt{n}}, \bar{X}+z_{\frac{\alpha}{2}} \frac{\alpha}{\sqrt{n}})
\label{conf}
\end{equation}
, where $\bar{X} \sim N(\mu, \frac{\sigma^2}{n})$, $n$ is the number of latent vectors in the green cluster, and $P(|Z| < z_{\frac{\alpha}{2}}) = 1 - \alpha$. Finally, we consider latent vectors smaller than $\hat{\theta_1}$ and larger than $\hat{\theta_2}$ as outlier vectors.

\subsubsection{Detection of C3 Words across Two Cybercrimes}
\label{sec:3.3.2}
If the green and red clusters are represented as sets $A$ and $B$, the output of the second method is a set of elements $x \in \{A \cap B\ \}$, where $x$ is a C3 word. This method is beneficial in analyzing mixed criminal activity involving drug use and sexual offences. 
Through our two-step approach, two dictionaries of drug and sex crime are constructed in advance. Then, given a document $d$ related to cybercrime as input, the ratio of drug C3 words and sex crime ones in $d$ is calculated as the output.

\subsubsection{Taxonomy of C3 Words}
\label{sec:3.3.3}
In general, C3 words relevant with drugs can be subdivided into c3 words such as marijuana, heroin, cocaine, etc. The same goes for sex crimes.
In this problem, given a set of C3 words (e.g., the cluster in green) as input, they are grouped to a set of clusters $\Phi = \{ \pi_1, \pi_2, ..., \pi_k \}$ through one of main clustering methods. Specifically, the number of clusters is approximately estimated using k-means++ in advance. Then, the k-means clustering method is performed. 
For each $\pi_i \in \Phi$, the C3 word corresponding to the centroid of $\pi_i$ is chosen as the C3 words representing $\pi_i$. We call it the category word. In addition,
if a C3 word is within $\epsilon$ from the centroid of $\pi_i$, where $\epsilon$ is a certain threshold value of the distance from the category, the C3 word is called a subcategory word. Finally, the output of the third method is a pair of the category word and four subcategory words per $\pi_i$.

\section{Experimental Results}
\label{expr}

To evaluate the proposed model, we used drug and sex crime related articles from Smart policing big data platform~\cite{smartpolicing} which is the same dataset used in~\citep{jin-etal-2023-darkbert}. We collected 360,116 drug and 249,970 sex crime related articles~\footnote{Drug crime data set: https://www.bigdata-policing.kr/product/view?product\_id=PRDT\_177. Sex crime data set: https://www.bigdata-policing.kr/product/view?product\_id=PRDT\_211.}. After removing the emoticons and special characters with numbers, we divide the data into train, validation and test as in 8:1:1 ratio. We designed the test data in three different ways to measure the model in different environments. First, we composed the drug and sex crime related articles as in 5:5 ratio to check the model can properly distinguish the articles in the Dark Web environment. Second, we added regular bulletin board articles from AIHub and mixed as in 6:2:2, 8:1:1, 9.9:0.05:0.05, and 99.99:0.005:0.005 ratios to check the model performance in Surface Internet environment.

We implemented eleven different models using python and PyTorch on Intel Core-i9 series CPU equipped with NVidia GeForce RTX3090 GPU and 250GB SSD.
At first, we implemented SAE, AAE, DAE, and VAE AutoEncoder based models, named Two-step(SAE), Two-step(AAE), Two-step(DAE), and Two-step(VAE), respectively. Then, we compared the performance with SDAE~\citep{cyberbullying2017}, GMM~\citep{GMMcyberbullying2020}, and MLM~\citep{euphemism2021}. The details of the parameters used in the experiments are shown in Table~\ref{table-env}.
To represent the traditional Word embedding based approach~\citep{ranaldi2022dark}, we build a GloVe model by training with our own data set, named dark-Glove. In addition, since DarkBERT~\citep{jin-etal-2023-darkbert} is not available in public, we used pretrained BERT-multi model after fine-tuned with our own data set, named dark-BERT. For dark-BERT, we used batch size 12 with one hidden layer by hardware memory limitation. We ran the model to 20 epochs with the parameters in Table~\ref{table-env}.

\begin{table}[tb]
\small
\centering
\begin{tabular}{cc}
\hline \hline
\textbf{Parameter} & \textbf{Description} \\
\hline \hline
Drop rate & 0.5  \\          
Batch size & 256 (16*) \\           
Learning rate & 0.0005 \\           
Epochs & 50 (20*) \\
Hidden layers & 2 \\ 
\hline \hline
\end{tabular}
\caption{\label{table-env} Experimental parameters. The dark-BERT used batch size 16 and 20 epochs by memory limitations.}
\end{table}

\begin{table}[tb]
\small
\centering
\begin{tabular}{cc}
\hline \hline
\textbf{Model} & \textbf{Error(\%)} \\
\hline \hline
SDAE~\citep{cyberbullying2017} & 5.2\%  \\
\hdashline
Two-step(SAE) & 2.7\% \\
Two-step(AAE) & 3.4\% \\
Two-step(DAE) & 5.1\% \\
Two-step(VAE) & 5.7\% \\
\hline \hline
\end{tabular}
\caption{\label{table0} Reconstruction Errors(\%).}
\end{table}

The first metric we used to measure the performance of the proposed models is a reconstruction error rate which is defined as a binary cross entropy of two classes between the original article and reconstructed version. Table~\ref{table0} shows the reconstruction error rates of five different AutoEncoder based models. The Two-step(SAE) shows the best performance as in 2.7\% reconstruction error rates and Two-step(VAE) shows the worst performance along with Two-step(DAE). Intuitively, VAE and DAE explores the variations of the given data during generating process which leads to low reconstruction rates.





\begin{table}[!ht]
\small
\centering
\begin{tabular}{cccc}
\hline \hline
\textbf{Model} & \textbf{Precision} & \textbf{Recall} &\textbf{F1-score}\\
\hline \hline
SDAE & \textbf{0.991} & 0.959 & 0.975 \\
GMM &  0.761 & 0.758 & 0.684 \\
MLM &  0.427 & 0.654 & 0.517 \\
dark-GloVe & 0.987 & 0.950 & 0.949 \\
dark-BERT & 0.903 & 0.955 & 0.925 \\
BERT-eup & 0.930 & 0.900 & 0.910 \\
LSHWE & 0.972 & 0.968 & 0.969 \\
\hdashline
Two-step(SAE) & \textbf{0.991} & 0.991 & \textbf{0.991} \\
Two-step(AAE) & 0.975 & \textbf{0.995} & 0.985 \\
Two-step(DAE) & \textbf{0.991} & 0.961 & 0.976 \\
Two-step(VAE) & 0.478 & 0.458 & 0.467 \\
\hline \hline
\end{tabular}
\caption{\label{f1score1}Precision, Recall and F1-score for each model with drug and sex crime related articles: SDAE~\citep{cyberbullying2017}, GMM~\citep{GMMcyberbullying2020}, MLM~\citep{euphemism2021}, dark-Glove~\citep{ranaldi2022dark}, dark-BERT~\citep{jin-etal-2023-darkbert}, BERT-eup~\citep{euphemism2021}, and LSHWE~\citep{LSHWE2020}.}
\end{table}

\begin{table}[!ht]
\small
\centering
\begin{tabular}{cccc}
\hline \hline
\textbf{Model} & \textbf{Precision} & \textbf{Recall} &\textbf{F1-score}\\
\hline \hline
SDAE & 0.737 & 0.621 & 0.673 \\
GMM  & 0.282 & 0.351 & 0.312 \\
MLM  & 0.065 & 0.333 & 0.111 \\
dark-GloVe & 0.414 & 0.663 & 0.464 \\
dark-BERT & 0.585 & 0.511 & 0.546 \\
BERT-eup & 0.416 & 0.683 & 0.517 \\
LSHWE & 0.506 & 0.488 & 0.497 \\
\hdashline
Two-step(SAE) & \textbf{0.756} & \textbf{0.718} & \textbf{0.737} \\
Two-step(AAE) & 0.692 & 0.646 & 0.666 \\
Two-step(DAE) & 0.705 & 0.693 & 0.699 \\
Two-step(VAE) & 0.334 & 0.110 & 0.165 \\
\hline \hline
\end{tabular}
\caption{\label{f1score2} Precision, Recall and F1-score for each model with general bulletin board articles with Drug and Sex crime related articles. The ratio for general bulletin board, drug and sex crime related articles as in 6:2:2.}
\end{table}

\begin{table}[!ht]
\small
\centering
\begin{tabular}{cccc}
\hline
\textbf{Model} & \textbf{Precision} & \textbf{Recall} &\textbf{F1-score}\\
\hline \hline
SDAE & 0.602 & 0.639 & 0.621 \\
GMM  & 0.189 & 0.244 & 0.213 \\
MLM  & 0.030 & 0.333 & 0.060 \\
dark-GloVe & 0.360 & 0.663 & 0.394 \\
dark-BERT & 0.557 & 0.590 & 0.572 \\
BERT-eup & 0.349 & 0.533 & 0.411 \\
LSHWE & 0.464 & 0.398 & 0.429 \\
\hdashline
Two-step(SAE) & \textbf{0.691} & \textbf{0.745} & \textbf{0.716} \\
Two-step(AAE) & 0.656 & 0.640 & 0.647 \\
Two-step(DAE) & 0.660 & 0.679 & 0.670 \\
Two-step(VAE) & 0.409 & 0.183 & 0.252 \\
\hline \hline
\end{tabular}
\caption{\label{f1score3} Precision, Recall and F1-score for each model with general bulletin board articles with Drug and Sex crime related articles. The ratio for general, drug and sex crime related articles as in 8:1:1.}
\end{table}

\begin{table}[!ht]
\small
\centering
\begin{tabular}{cccc}
\hline
\textbf{Model} & \textbf{Precision} & \textbf{Recall} &\textbf{F1-score}\\
\hline \hline
Case 1            & 0.756   & 0.718 & 0.737 \\
Case 2            & 0.691   & 0.745 & 0.716 \\
Case 3            & 0.602   & 0.660 & 0.631 \\
Case 4            & 0.667   & 0.793 & 0.716 \\
\hline \hline
\end{tabular}
\caption{\label{f1score4} Precision, Recall and F1-score of Two-step(SAE) for various settings with general bulletin board articles with Drug and Sex crime related articles. The ratio for general, drug and sex crime related articles as in Case 1(6:2:2), Case 2(8:1:1), Case 3(9.9:0.05:0.05), Case 4(99.99:0.005:0.005).}
\end{table}

Next, we measured the precision, recall and F1-score as shown in Table~\ref{f1score1}. The Two-step(SAE) approach shows the superior performance with 0.991 in F1-score compared to all others. All other Two-step with AutoEncoder based models show the better performance than others except the Two-step(VAE). Considering the input sentence does not have any context information, the lower performance of GMM and MLM are as expected. 

To test with Surface Internet environment, we added regular bulletin board articles with other cybercrime related articles. Table~\ref{f1score2} and~\ref{f1score3} shows the results in different ratio mixes. For the regular, drug and sex crime related articles as in 6:2:2 ratio, Two-step(SAE) and Two-step(DAE) shows the best performance as shown in Table~\ref{f1score2}. Interestingly, the dark-GloVe shows the better performance than dark-BERT which was as indicated in~\citep{ranaldi2022dark}. However, dark-BERT starts to get better with general bulletin board articles as in Figure~\ref{f1score2}, and surpasses the dark-GloVe with more general bulletin board articles as in Figure~\ref{f1score3}. We conjecture that dark-BERT represents the semantic context better than domain specifically trained dark-GloVe with more general bulletin board articles(i.e., 8:1:1 ratio). 
To check the performance of Two-step(SAE) in more realistic scenarios, we experimented in various mixtures of articles. Table~\ref{f1score4} shows the experimental results. Even in the extreme case (99.99\% general articles with 0.005 drug and sex crime related articles), the Two-step(SAE) shows a decent performance. 

\begin{figure*}[!ht]
\begin{center}
\includegraphics[scale=0.8]{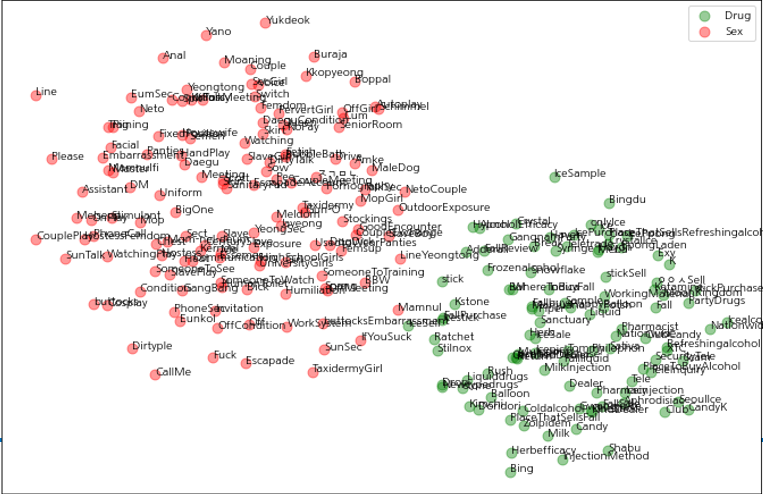} 
\caption{C3 words related to drug and sex crimes.}
\label{fig.5}
\end{center}
\end{figure*}

{\bf Recently added C3 words:} We assume that new C3 words can be found as outliers in the latest articles (Section~\ref{sec:3.3.1}). To prove the latter, we collected recently added C3 words using our framework (i.e., dated as in 12.31.2022) and confirmed that 93.6\% out of 756 for drug and 86.6\% out of 679 for sex crime related words are matching as the outliers. For example, the `ICE' and `stilnox' are found as example of outliers in drug related articles and `slave' and `fetish' are found as in sex crime related outliers from the dated articles. We can add those in the C3 word dictionary unless they are already in the list.

{\bf Overlapped C3 words:} We assume that a considerable amount of words are related with multiple crimes (Section~\ref{sec:3.3.2}).
The 1,025 articles out of 36,000 articles (i.e., 2.85\%) are overlapped in two different crimes, and 1,476 unique words were found from those articles. For example, `Ecstasy' and `Adellerall' are found in the drug and sex crime related articles at the same time. Those are known as drugs used to intoxicate women before sex crime.

\begin{figure}[!ht]
\begin{center}
\includegraphics[scale=0.6]{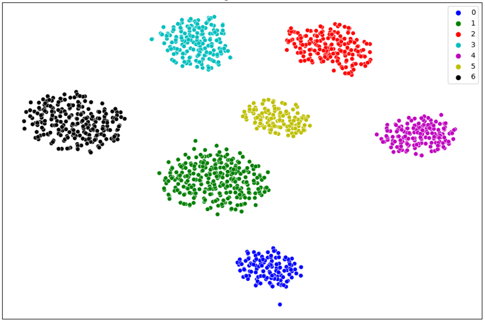} 
\includegraphics[scale=0.585]{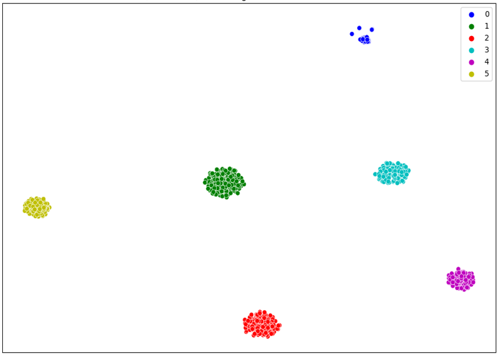} 
\caption{Categories of drug (top) and sex crime (bottom) related words.}
\label{fig.6}
\end{center}
\end{figure}

{\bf C3 word taxonomy: } Figure~\ref{fig.5} shows the drug and sex crime related word lists. It shows that drug and sex crime related words are close to each other with some overlap. To explore further, we applied k-means clustering on the C3 words. Figure~\ref{fig.6} shows the subgroups of the drug and sex crime related words. The drug related words are categorized into seven subgroups and each subgroup has some associated word lists. Sex crime related words show the similar results with six subgroups. Through this process, we can categorize crime related words into taxonomy (Section~\ref{sec:3.3.3}). Table~\ref{table-taxonomy} shows the detail lists of the categories and subcategories of the C3 words for drug and sex crime related articles.

\begin{table}[!ht]
\scriptsize
\centering
\begin{tabular}{|c|c|l|}
\hline
 & \textbf{Category} & \textbf{Subcategory} \\
\hline
Drug & Sleeping Aid & 0: zolpidem, stilnox, rohypnol, buy  \\
     & ICE          & 1: chansul, hwinsul, bingdu, icekeki \\
     & Stimulant    & 2: methamphetamine, taiproduct, meth \\
     & Trade        & 3: FCFS, confirm, pickup, drop \\
     & Liquid Drop  & 4: hub, Candy, hubLiquid, ecstasy \\
     & Mulppoing    & 5: sedation, anesthesia, cannabis \\
     & Rush         & 6: rushHopper, adderalll, propofol \\
\hline
Sex    & Teleline   & 0: event, invitation, money, 3Ps \\
       & Courgar    & 1: pussy, doll, violate, married \\
       & Off        & 2: pay, squirt, OffMeet, petting \\
       & Master     & 3: SM, slave, violate, horny     \\ 
       & Neto       & 4: couple, meet, kakao, netocouple \\
       & Bondage    & 5: feti, fetish, rape, whore \\
\hline
\end{tabular}
\caption{\label{table-taxonomy} Categorized C3 words list.}
\end{table}

\section{Conclusion}
\label{cond}
In this paper, we proposed a two-step cybercrime coded word detection method using multi-level representation learning. Two-step(SAE) showed 0.991 in F-score which is higher than those of dark-GloVe and dark-BERT models. In addition, our proposed model detects new words recently coined through outliers, and words related with multiple crimes. At the end, we can build and organize the collected C3 word lists into categories and subcategories. 

Our experiments are limited into two different cybercrimes namely drug and sex crime related. In a real environment, there are multiple cybercrimes (i.e., cyberbullying, drug, sex crime, weapons, etc.) in multiple social media platforms. Extending our model to support multiple crimes on various social media environments will be our future works.

\section*{Limitations}
The proposed approach has limitations inherited from the training data set. Since the training data set size is relatively small, the performance is limited by the data size. Considering the false positive could severely damage the fame of individual or company, it should be carefully applied in real practice. The second limitation is based on the regional bias or specification. Since the model is trained by the korean police defined cyber-crime data, the model should go through regional adapation since some of the drugs or activities might be legal in other countries.

\section*{Ethics Statement}
The major concerns of the proposed approach might be an abusing of the technology to monitor or detect dissidents by government administration. In addition, it could be misused to avoid police watch by criminal group. At the same time, the technology also generates a privacy concern. Depersonalization should be properly applied to avoid the potential privacy issue. The last issue might come from false positive. Since the falsely accusing could severely defamate a specific personnel, group, or company, the technology should be carefully applied to avoid the disaster.

\section*{Acknowledgement}
This work was supported by the National Research Foundation of Korea (NRF) funded by the Korean Government through MSIT under Grant NRF-2022R1A2C1011404. The authors would like to thank the anonymous reviewers for their feedback.

\section*{Bibliographical References}\label{sec:reference}

\bibliographystyle{lrec-coling2024-natbib}
\bibliography{lrec-coling2024-example}

\begin{thebibliography}{18}
\expandafter\ifx\csname natexlab\endcsname\relax\def\natexlab#1{#1}\fi

\bibitem[{Agency(2022{\natexlab{a}})}]{police-22}
Korean National~Police Agency. 2022{\natexlab{a}}.
\newblock Korean national police agency.
\newblock In \emph{https://www.police.go.kr/index.do}.

\bibitem[{Agency(2022{\natexlab{b}})}]{smartpolicing}
Korean National~Police Agency. 2022{\natexlab{b}}.
\newblock Smart policing big data platform.
\newblock \url{https://www.bigdata-policing.kr}.
\newblock Accessed: 2023-10-310.

\bibitem[{Al~Nabki et~al.(2017)Al~Nabki, Fidalgo, Alegre, and
  de~Paz}]{al-nabki-etal-2017-classifying}
Mhd~Wesam Al~Nabki, Eduardo Fidalgo, Enrique Alegre, and Ivan de~Paz. 2017.
\newblock \href {https://aclanthology.org/E17-1004} {Classifying illegal
  activities on tor network based on web textual contents}.
\newblock In \emph{Proceedings of the 15th Conference of the {E}uropean Chapter
  of the Association for Computational Linguistics: Volume 1, Long Papers},
  pages 35--43, Valencia, Spain. Association for Computational Linguistics.

\bibitem[{Al~Nabki et~al.(2019)Al~Nabki, Fidalgo, Alegre, and
  Fernández-Robles}]{al-nabki-etal-2019}
Wesam Al~Nabki, Eduardo Fidalgo, Enrique Alegre, and Laura Fernández-Robles.
  2019.
\newblock \href {https://doi.org/10.1016/j.eswa.2019.01.029} {Torank:
  Identifying the most influential suspicious domains in the tor network}.
\newblock \emph{Expert Systems with Applications}, 123.

\bibitem[{Avarikioti et~al.(2018)Avarikioti, Brunner, Kiayias, Wattenhofer, and
  Zindros}]{Avarikioti2018StructureAC}
Georgia Avarikioti, Roman Brunner, Aggelos Kiayias, Roger Wattenhofer, and
  Dionysis Zindros. 2018.
\newblock \href {https://api.semanticscholar.org/CorpusID:53222189} {Structure
  and content of the visible darknet}.
\newblock \emph{ArXiv}, abs/1811.01348.

\bibitem[{Cheng et~al.(2020)Cheng, Shu, Wu, Silva, Hall, and
  Liu}]{GMMcyberbullying2020}
Lu~Cheng, Kai Shu, Siqi Wu, Yasin~N. Silva, Deborah~L. Hall, and Huan Liu.
  2020.
\newblock \href {http://arxiv.org/abs/2008.02642} {Unsupervised cyberbullying
  detection via time-informed gaussian mixture model}.
\newblock \emph{CoRR}, abs/2008.02642.

\bibitem[{Choshen et~al.(2019)Choshen, Eldad, Hershcovich, Sulem, and
  Abend}]{choshen-etal-2019-language}
Leshem Choshen, Dan Eldad, Daniel Hershcovich, Elior Sulem, and Omri Abend.
  2019.
\newblock \href {https://doi.org/10.18653/v1/P19-1419} {The language of legal
  and illegal activity on the {D}arknet}.
\newblock In \emph{Proceedings of the 57th Annual Meeting of the Association
  for Computational Linguistics}, pages 4271--4279, Florence, Italy.
  Association for Computational Linguistics.

\bibitem[{Devlin et~al.(2019)Devlin, Chang, Lee, and
  Toutanova}]{devlin-etal-2019-bert}
Jacob Devlin, Ming-Wei Chang, Kenton Lee, and Kristina Toutanova. 2019.
\newblock \href {https://doi.org/10.18653/v1/N19-1423} {{BERT}: Pre-training of
  deep bidirectional transformers for language understanding}.
\newblock In \emph{Proceedings of the 2019 Conference of the North {A}merican
  Chapter of the Association for Computational Linguistics: Human Language
  Technologies, Volume 1 (Long and Short Papers)}, pages 4171--4186,
  Minneapolis, Minnesota. Association for Computational Linguistics.

\bibitem[{Jin et~al.(2023)Jin, Jang, Cui, Chung, Lee, and
  Shin}]{jin-etal-2023-darkbert}
Youngjin Jin, Eugene Jang, Jian Cui, Jin-Woo Chung, Yongjae Lee, and Seungwon
  Shin. 2023.
\newblock \href {https://doi.org/10.18653/v1/2023.acl-long.415} {{D}ark{BERT}:
  A language model for the dark side of the {I}nternet}.
\newblock In \emph{Proceedings of the 61st Annual Meeting of the Association
  for Computational Linguistics (Volume 1: Long Papers)}, pages 7515--7533,
  Toronto, Canada. Association for Computational Linguistics.

\bibitem[{Jin et~al.(2022)Jin, Jang, Lee, Shin, and
  Chung}]{jin-etal-2022-shedding}
Youngjin Jin, Eugene Jang, Yongjae Lee, Seungwon Shin, and Jin-Woo Chung. 2022.
\newblock \href {https://doi.org/10.18653/v1/2022.naacl-main.412} {Shedding new
  light on the language of the dark web}.
\newblock In \emph{Proceedings of the 2022 Conference of the North American
  Chapter of the Association for Computational Linguistics: Human Language
  Technologies}, pages 5621--5637, Seattle, United States. Association for
  Computational Linguistics.

\bibitem[{Kim(2022)}]{bkim-22}
B.~Kim. 2022.
\newblock Drug crime white paper.
\newblock In \emph{Supreme (Public) Prosecutors' Office}.

\bibitem[{Latapy et~al.(2012)Latapy, Magnien, and
  Fournier-S’niehotta}]{Latapy2012QuantifyingPA}
Matthieu Latapy, Cl{\'e}mence Magnien, and Rapha{\"e}l Fournier-S’niehotta.
  2012.
\newblock \href {https://api.semanticscholar.org/CorpusID:1919337} {Quantifying
  paedophile activity in a large p2p system}.
\newblock \emph{ArXiv}, abs/1206.4166.

\bibitem[{Morris and Hirst(2012)}]{Morris2012IdentifyingSP}
Colin Morris and Graeme Hirst. 2012.
\newblock \href {https://api.semanticscholar.org/CorpusID:14777163}
  {Identifying sexual predators by svm classification with lexical and
  behavioral features}.
\newblock In \emph{Conference and Labs of the Evaluation Forum}.

\bibitem[{Pennington et~al.(2014)Pennington, Socher, and
  Manning}]{pennington-etal-2014-glove}
Jeffrey Pennington, Richard Socher, and Christopher Manning. 2014.
\newblock \href {https://doi.org/10.3115/v1/D14-1162} {{G}lo{V}e: Global
  vectors for word representation}.
\newblock In \emph{Proceedings of the 2014 Conference on Empirical Methods in
  Natural Language Processing ({EMNLP})}, pages 1532--1543, Doha, Qatar.
  Association for Computational Linguistics.

\bibitem[{Ranaldi et~al.(2022)Ranaldi, Nourbakhsh, Patrizi, Ruzzetti, Onorati,
  Fallucchi, and Zanzotto}]{ranaldi2022dark}
Leonardo Ranaldi, Aria Nourbakhsh, Arianna Patrizi, Elena~Sofia Ruzzetti, Dario
  Onorati, Francesca Fallucchi, and Fabio~Massimo Zanzotto. 2022.
\newblock \href {http://arxiv.org/abs/2201.05613} {The dark side of the
  language: Pre-trained transformers in the darknet}.

\bibitem[{Zhao and Mao(2017)}]{cyberbullying2017}
Rui Zhao and Kezhi Mao. 2017.
\newblock \href {https://doi.org/10.1109/TAFFC.2016.2531682} {Cyberbullying
  detection based on semantic-enhanced marginalized denoising auto-encoder}.
\newblock \emph{IEEE Transactions on Affective Computing}, 8(3):328--339.

\bibitem[{Zhao et~al.(2020)Zhao, Gao, Luo, Zhang, and Xiong}]{LSHWE2020}
Zehua Zhao, Min Gao, Fengji Luo, Yi~Zhang, and Qingyu Xiong. 2020.
\newblock \href {https://doi.org/10.1109/IJCNN48605.2020.9207640} {Lshwe:
  Improving similarity-based word embedding with locality sensitive hashing for
  cyberbullying detection}.
\newblock In \emph{2020 International Joint Conference on Neural Networks
  (IJCNN)}, pages 1--8.

\bibitem[{Zhu et~al.(2021)Zhu, Gong, Bansal, Weinberg, Christin, Fanti, and
  Bhat}]{euphemism2021}
Wanzheng Zhu, Hongyu Gong, Rohan Bansal, Zachary Weinberg, Nicolas Christin,
  Giulia Fanti, and Suma Bhat. 2021.
\newblock \href {https://doi.org/10.1109/SP40001.2021.00075} {Self-supervised
  euphemism detection and identification for content moderation}.
\newblock In \emph{Proceedings - 2021 IEEE Symposium on Security and Privacy,
  SP 2021}, Proceedings - IEEE Symposium on Security and Privacy, pages
  229--246, United States. Institute of Electrical and Electronics Engineers
  Inc.

\end{thebibliography}


\appendix
\section{Appendices}

\subsection{Data Example}
Table~\ref{table-input} shows a sample input data collected from the Smart Policing Big Data Platform Network~\cite{smartpolicing}. As shown in the example, the sentence is composed of broken C3 words and context information is not available from the given sentence. The latter makes it difficult to apply a conventional language model relying on the context information in the sentences and document including word embedding and pre-trained large language model. As shown in the experimental results, GloVe and BERT based approaches have limitations by the aforementioned characteristics of the data.

\begin{table}[h]
\scriptsize
\centering
\begin{tabular}{|c|m{2.5cm}|m{2.5cm}|}
\hline
 & \textbf{Original} & \textbf{Partially translated} \\
\hline
Drug & \raisebox{-\totalheight}{\includegraphics[width=0.15\textwidth, height=20mm]{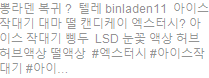}}
     & Back BbongRaden? Tele binladen11 Icekeki Weed Exstasy? Bbongo LSD SnowWhite Liquid Hub HibLiquid \#Ecstasy \#Icekeki \#ICE \\
\hline
Sex  & \raisebox{-\totalheight}{\includegraphics[width=0.15\textwidth, height=20mm]{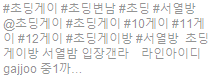}}
     & \#gay \#purvert \#underagepurvert \#order \#underageguy \#10gay allnude lineID gajoo to14\\
\hline
\end{tabular}
\caption{\label{table-input} Data samples.}
\end{table}

\subsection{Newly added and overlapped C3 word supplements}
\begin{figure}[!ht]
\begin{center}
\includegraphics[scale=0.33]{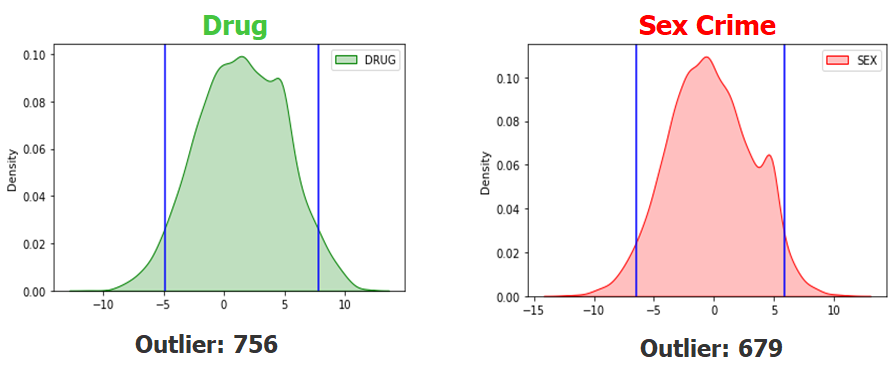} 
\includegraphics[scale=0.33]{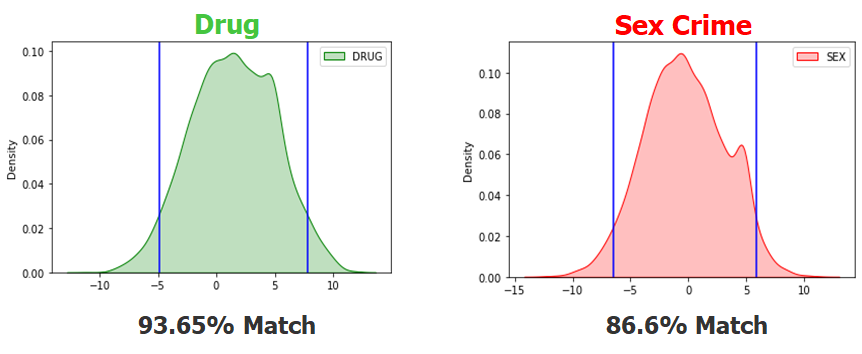} 
\caption{Outlier C3 word: newly added words for drug and sex crime related are outliers.}
\label{fig.3}
\end{center}
\end{figure}

\begin{figure}[!ht]
\begin{center}
\includegraphics[scale=0.3]{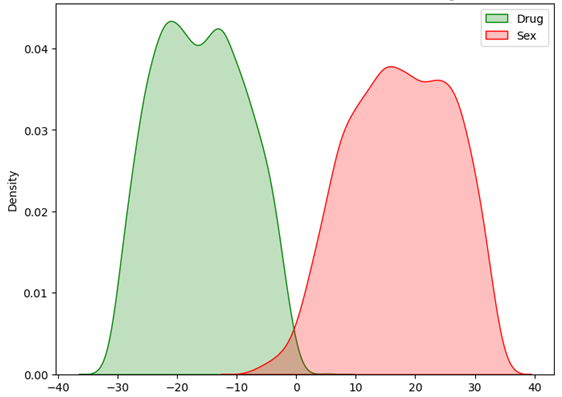} 
\includegraphics[scale=0.4]{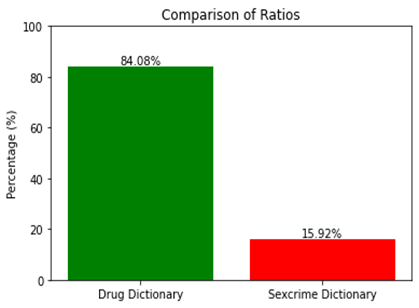} 
\caption{Overlapped in drug and sex crime related words.}
\label{fig.4}
\end{center}
\end{figure}

Figure~\ref{fig.3} shows that the recently added C3 words which are located as outliers in the recent articles. The total number of C3 words are 756 and 679 for drug and sex crime related articles, respectively. We can find 708 (93.65\%) and 588 (86.6\%) C3 words from the recent article (dated as in 12.31.2022) among the list. Table~\ref{table-outliers} shows the detail information for the outliers of the latest articles.
Considering some of the drugs are used for sex crime purpose, we assume that many euphemism words are related in multiple crimes. Figure~\ref{fig.4} shows the overlapped region in drug and sex crime related articles. Many words are used in multiple crimes, and the examples are shown in Table~\ref{table-overlap}.

\begin{table}[!ht]
\scriptsize
\centering
\begin{tabular}{|c|m{2.5cm}|m{2.5cm}|}
\hline
 & \textbf{Original} & \textbf{Partially translated} \\
\hline
Drug & \raisebox{-\totalheight}{\includegraphics[width=0.1\textwidth, height=20mm]{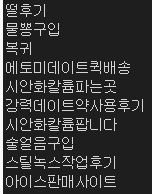}}
     &  ICE stilnox \\
\hline
Sex & \raisebox{-\totalheight}{\includegraphics[width=0.06\textwidth, height=20mm]{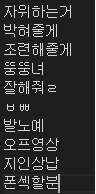}}
     &  off slave phonesex \\
\hline
\end{tabular}
\caption{\label{table-outliers} Outlier words.}
\end{table}

\begin{table}[!ht]
\scriptsize
\centering
\begin{tabular}{|c|m{2.5cm}|m{2.5cm}|}
\hline
 & \textbf{Original} & \textbf{Partially translated} \\
\hline
Drug and Sex & \raisebox{-\totalheight}{\includegraphics[width=0.15\textwidth, height=25mm]{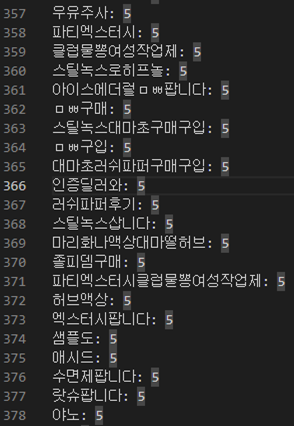}}
     &  adderall ecstasy zolpidem escort \\
\hline
\end{tabular}
\caption{\label{table-overlap} Overlapped words.}
\end{table}


\end{document}